# Generating Findings for Jaw Cysts in Dental Panoramic Radiographs Using a GPT-Based VLM: A Preliminary Study on Building a Two-Stage Self-Correction Loop with Structured Output (SLSO) Framework

Nanaka Hosokawa[1], Ryo Takahashi[2], Tomoya Kitano[3], Yukihiro Iida[3], Chisako Muramatsu[4], Tatsuro Hayashi[2], Yuta Seino[5], Xiangrong Zhou[6], Takeshi Hara[6], Akitoshi Katsumata[3], and Hiroshi Fujita[6]

[1] Graduate School of Natural Science and Technology, Gifu University, Gifu, Japan
[2] EyeTech Co., Ltd, Tokyo, Japan
[3] School of Dentistry, Asahi University, Mizuho, Japan
[4] Faculty of Data Science, Shiga University, Hikone, Japan
[5] Division for Oral Dental Informatics, The University of Osaka Dental Hospital, Osaka, Japan
[6] Faculty of Engineering, Gifu University, Gifu, Japan

## Abstract

Vision-language models (VLMs) such as GPT (Generative PreTrained Transformer) have shown potential for medical image interpretation; however, challenges remain in generating reliable radiological findings in clinical practice, as exemplified by dental pathologies. This study proposes a Self-correction Loop with Structured Output (SLSO) framework as an integrated processing methodology to enhance the accuracy and reliability of AI-generated findings for jaw cysts in dental panoramic radiographs.

Dental panoramic radiographs with jaw cysts were used to implement a 10-step integrated processing framework incorporating image analysis, structured data generation, tooth number extraction, consistency checking, and iterative regeneration. The framework functioned as an external validation mechanism for GPT outputs. Performance was compared against the conventional Chain-of-Thought (CoT) method across seven evaluation items: transparency, internal structure, borders, root resorption, tooth displacement, relationships with other structures, and tooth number.

The SLSO framework improved output accuracy for multiple items compared to the CoT method, with the most notable improvements observed in tooth number identification, tooth displacement detection, and root resorption assessment. In successful cases, consistently structured outputs were achieved after up to five regenerations. The framework enforced explicit negative finding descriptions and suppressed hallucinations, although accurate identification of extensive lesions spanning multiple teeth remained limited.

This investigation established the feasibility of the proposed integrated processing methodology and provided a foundation for future validation studies with larger, more diverse datasets.

## 1. Introduction

In recent years, large language models (LLMs) such as OpenAI's ChatGPT have rapidly expanded the application of natural language processing in the medical field. They have been employed in tasks such as medical record summarization, dialog support, and the automatic generation of radiological findings. With the emergence of GPT-4o, interest in multimodal diagnostic support has grown, particularly in its ability to integrate visual and textual information. For example, in a meta-analysis of 330 studies, Ye et al. concluded that multimodal LLMs, including GPT-4o, have been widely applied for medical report generation and diagnostic support, with notable progress in visual-text integration[1]. Furthermore, a recent scoping review by Zhou et al. found that vision-language models (VLMs) such as GPT-4V, LLaVA, and Flamingo have been applied to diagnostic tasks by combining radiological images with clinical text, and several studies have reported encouraging results[2]. By contrast, the application of LLMs and VLMs in dentistry remains in its infancy. Practical investigations using GPT-4o have only emerged since late 2024. On text-based assessments, GPT-4-based models have shown promising results. For example, GPT-4o outperformed GPT-4 on text-based questions from Japan's 117th National Dental Examination and achieved high performance in academic written examinations, in some cases exceeding the average scores of dental students[3,4]. Jaworski et al. reported that GPT-4o achieved 70.85% overall accuracy on 200 text-based questions from the Polish National Dental Examination, though performance was much lower (36.36%) on clinical case-based items compared to general knowledge questions (72.87%)[5]. However, when GPT-4V was evaluated on image-based questions from the 116th Japanese National Dental Examination— including intraoral and extraoral photographs, panoramic and dental radiographs, CT, MRI, and pathological images—the overall correct response rate was only 35.0%, with particularly low performance (28.6%) on clinical practical questions requiring integration of multiple imaging modalities[6].

These studies indicate that while current LLMs show promise in text-based dental examinations, VLMs still lack the visual recognition accuracy required for reliable dental image interpretation. Recently, several multimodal dental imaging applications have been explored. For instance, Dasanayaka et al. proposed methods using GPT-4o for generating findings from panoramic radiographs, showing potential in detecting caries and impacted teeth[7]. Several studies have explored VLM-based approaches for dental imaging tasks such as detection or classification of dental conditions[8–11]. However, these studies highlight the limitations of general-purpose models in visual inference. In the specific domain of cyst diagnosis in panoramic radiographs, which is the focus of this study, prior work has been limited to text-based approaches, such as Silva et al.'s study using GPT-3.5 with textual clinical inputs[12]. While Dasanayaka et al.[7] explored finding generation for panoramic radiographs, their study targeted different pathologies (caries and impacted teeth) from those addressed in the present study. Other studies such as those by Pham[8], A¸sar et al.[9], Liu et al.[10], and Camlet et al.[11] have primarily focused on detection or classification of dental conditions using VLMs. This represents a significant gap and challenge in the current literature. Another challenge in VLMs lies in hallucinations and factual inconsistencies, which remain critical concerns for medical AI applications. For example, Wang et al. conducted a comprehensive investigation into the factuality of LLMs and concluded that strategies such as Chain-of-Thought (CoT) prompting and retrieval-augmented generation (RAG) contribute to suppressing hallucinations, highlighting the importance of reliability in specialized medical domains[13]. Alkaissi and McFarlane documented erroneous citations and fabricated facts generated by ChatGPT (version not specified in the original study), underscoring the need for careful verification in healthcare contexts[14]. Chang et al. recently introduced the "MedHEval" benchmark for systematic evaluation and mitigation of hallucination in multimodal medical models, emphasizing that robust output-verification mechanisms are essential for safe deployment of VLMs[15]. As such, although GPT-based models have demonstrated a certain level of effectiveness in dental diagnostic support, they still face limitations in ensuring spatial consistency, factual accuracy, and comprehensive domain knowledge, and a substantial risk of erroneous outputs remains. Therefore, to enhance the reliability and practicality of VLMs in dentistry, it is necessary to establish a framework that integrates consistency checks with structured information, visual data preprocessing, and hallucination-suppression techniques. In this preliminary study, we developed an integrated framework that functions as an external validation mechanism by harnessing the multimodal capabilities of a GPT-based VLM to support the interpretation of cysts in dental panoramic radiographs. This framework, called Self-correction Loop with Structured

Output (SLSO), incorporates both structured-data generation and finding-text generation. The aim of this study is to investigate the feasibility of the proposed SLSO framework as a structured self-correction mechanism designed to improve the consistency and reliability of AI-generated radiological findings.

# 2. Materials and Methods

## 2.1 Pilot Experiments

As a preliminary investigation, a series of pilot experiments were conducted to refine the system design. Four phases were examined: (1) direct finding generation from resized panoramic images, (2) manual ROI (Region of Interest) extraction around cysts, (3) incorporation of tooth segmentation and tooth number annotation, and (4) introduction of CoT prompts. These preliminary studies revealed major limitations, including difficulties in tooth number identification, frequent vague expressions, and hallucinations. The insights gained from these investigations motivated the design of the proposed SLSO framework with a self-correction loop, described in detail in the following section. Detailed procedures, example outputs, and phase-specific results are provided in Supplementary Note S1 of the Supplementary Information.

## 2.2 Dataset and Ground-Truth Creation

For this study, dental panoramic radiographs of 22 jaw cysts (cases_001–022) were collected at the Asahi University Medical and Dental Center (Mizuho, Gifu, Japan). Each image was annotated with the jaw cyst and tooth margins as well as the Fédération Dentaire Internationale (FDI) number for each tooth. A dental radiologist (T.K.) manually performed all annotations. None of the cases included in this study were histopathologically confirmed. Instead, all diagnoses were established based on radiographic findings and clinical assessment by an experienced oral and maxillofacial radiologist. For this preliminary study, ground truth annotations were created by a single expert reviewer (T.K., an experienced oral and maxillofacial radiologist). Future studies involving larger and more diverse datasets will implement multi-rater consensus approaches to ensure robust reference standards. The corresponding ground-truth findings were created by the same radiologist and saved as text files. The number of teeth affected by jaw cysts (number of affected teeth) ranged from one to six, with an average of 2.8. For each case, the ground-truth findings were standardized to describe the lesion location and extent in the format "from tooth # to tooth #," and to include the following three mandatory radiographic features: radiolucency, margination, and multilocularity. Furthermore, anatomical effects such as root resorption, tooth displacement, and their relationship with the cortical bone were also described, and the names of likely diseases in the differential diagnosis were stated when necessary. The minimum number of words in the ground-truth findings was 68, the maximum was 201, and the average was 127.3 ± 31.2 words. All patient data were collected and de-identified at Asahi University with the approval of the Institutional Review Board (Approval No. 32040). Subsequent data analysis was conducted at Gifu University with the approval of the Institutional Review Board (Approval No. 2020-250). All the procedures complied with the principles of the Declaration of Helsinki.

### 2.3 Structured Output Schema and Image Annotation

Free-form written findings often contain variability and ambiguity, with synonyms such as "clear border," "clear margin," and "sharp contour" used interchangeably, and descriptions are prone to omission or hallucination (the inclusion of nonexistent findings). To mitigate these issues, we adopted a structured approach that constrains outputs to predefined options, thereby reducing variability, enabling machine-readable knowledge extraction, enhancing verifiability, and accumulating reusable structured data. Guided by collaboration with a co-author dental radiologist, we defined an explicit schema for the interpretation of items related to jaw cysts. In this study, a schema refers to a structured template that specifies the organization of interpretation categories and constraints on allowable outputs, ensuring consistency and reproducibility. Using this scheme, we classified the interpretation items into seven categories and assigned predefined options with labels, as summarized in Table 1. Notably, the three-level categorization (no/mild/severe') for root resorption and tooth displacement was adopted to provide a clinically intuitive and reproducible framework, reflecting commonly used radiological assessment practices while maintaining simplicity for structured evaluation.

**Table 1** Structured category and output values based on schema design

| Category | Output Value |
|---|---|
| **X-ray transparency** | radiolucent |
| | radiopaque |
| **Internal structure** | unilocular |
| | multilocular |
| **Border** | well-defined |
| | ill-defined |
| **Root resorption** | no |
| | mild |
| | severe |
| **Tooth displacement** | no |
| | mild |
| | severe |
| **Relationship to mandibular canal / maxillary sinus** | contact |
| | displacement |
| | invasion |
| **Affected tooth number** | 11, 12, 13, ...* |

* FDI notation is used for tooth numbering.

For the image input, we employed annotated ROI images, including tooth margins and tooth numbers, produced as described in "Supplementary Note S1 of the Supplementary Information" (see Phase 3 and Figure S1d for details). These annotations were designed to guide the model in recognizing tooth boundaries and numbers while also providing standardized inputs that ensured consistency between schema-based interpretation and visual data. Representative examples of annotated input images used in this process are shown in Supplementary Figure S1.

### 2.4 Overall Flow of Proposed Framework

The image interpretation support framework developed in this study used dental panoramic radiographs as the input, generates structured data and natural language findings regarding jaw cysts in a step-by-step manner, and includes a process for evaluating and correcting the consistency of each output (see Figure 1 for an overview of the processing flow). A brief outline of the ten sequential steps is provided here, and detailed descriptions, prompt examples, and schema definitions are available in Supplementary Note S2 of Supplementary Information.

1. **Input:** ROI images with annotated tooth margins and numbers, together with interpretation instructions, are provided.
2. **GPT-4o Image Analysis:** Multimodal analysis of the ROI image is initiated.
3. **Structured Data Generation:** Schema-based structured outputs are generated in JSON format.
4. **Tooth Number Extraction:** Affected tooth numbers are extracted directly from the image.
5. **Tooth Number Consistency Check:** Structured data and extracted tooth numbers are compared for consistency.
6. **Regenerate Structured Data & Re-extract Tooth Numbers:** Structured data and tooth numbers are regenerated when mismatches are detected.
7. **Finding Generation:** Radiology findings in natural language are generated from the structured data.
8. **Regenerate Structured Data from Findings:** Generated findings are converted back into structured data for verification.
9. **Structured Data Consistency Check:** Consistency between restructured and original structured data is checked.
10. **Regenerate Findings:** Findings are regenerated if inconsistencies remain, yielding the final output.

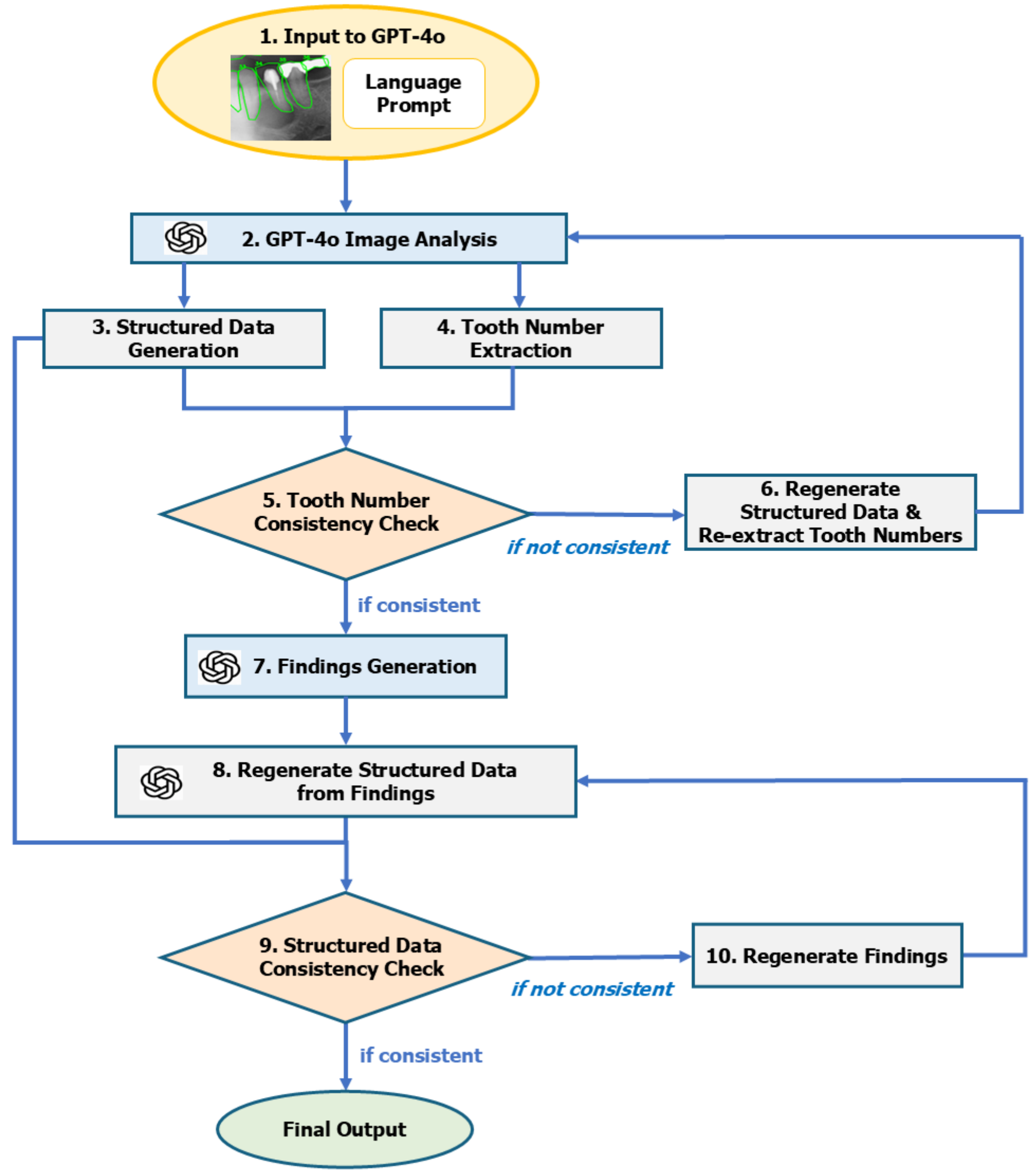

**Figure 1** Overview of proposed Self-correction Loop with Structured Output (SLSO) framework.

### 2.5 Experimental Setup

We compared the proposed SLSO framework with the conventional CoT method in terms of accuracy and consistency of the radiological findings generated for dental panoramic radiographs. To clarify the statistical evaluation framework, we defined the following null hypothesis: H0: The application of the proposed SLSO framework does not improve the diagnostic performance of the GPT-based VLM compared with the conventional CoT approach. Because this study focuses primarily on methodological feasibility using a limited dataset, this statistical formulation was introduced to support comparative evaluation rather than to perform a definitive hypothesis-driven validation.

### 2.5.1 Evaluation Criteria

To evaluate the accuracy of the generated findings, all structured items defined in the schema (e.g., location, transparency, internal structure, border, and affected teeth) were used as evaluation criteria. Each item was assigned a score of 0/1 (perfect match or not) for quantitative evaluation.

Score 1: Semantically consistent with the ground-truth findings (same meaning, no discrepancies in description).

Score 0: Inconsistent or missing (misinterpretation of meaning, omission, incorrect description, etc.).

This scoring allowed us to evaluate the accuracy of each schema item for each case and compare the overall performance by calculating the average accuracy. Although this binary exact-match scoring ensured objectivity and reproducibility, it may underestimate outputs that are partially correct or semantically close to the ground truth. More fine-grained evaluation methods such as semantic similarity metrics or expert-based rating scales should be incorporated into future studies to better capture its practical clinical utility.

### 2.5.2 Evaluation Procedure

Both methods (proposed SLSO framework and direct generation using CoT) were applied to all 22 cases. The score was calculated for each structured item in the generated output. Additionally, a case-based analysis of representative cases was conducted to examine whether the consistency check and regeneration process contributed to the suppression of incorrect descriptions and hallucinations.

### 2.5.3 Model and API Configuration

The model used in this study was the GPT-4o (gpt-4o-2024-11-20), a multimodal LLM provided by OpenAI. It was adopted because it supports both image and text input and can integrate image interpretation and natural language generation for dental panoramic radiographs.

For inference using the API, a Base64-encoded image together with a text prompt was simultaneously input via the gpt-4o endpoint. The output was structured data in JSON format or radiological findings, depending on the prompt design.

To stabilize the behavior of the model and maintain consistent outputs from the same image, the following parameter settings were used:

- temperature: 0.2 (low to increase output certainty)
- top_p: 1.0 (considering all top values in the probability distribution)
- max_tokens: 2048 (maximum tokens generated)
- frequency_penalty: 0.0 (no suppression of word repetition)
- presence_penalty: 0.0 (no induction of new words)

These settings minimize output fluctuations while achieving high reproducibility and accuracy for structured data and sentences.

#### 2.5.4 Prompt Design

In this study, multiple prompt designs were used depending on the task to improve the accuracy and consistency of finding generation.

**1) Prompts for Structured Data Generation**

In the structured output-based approach, we predefined a schema containing the interpretation items required for radiological diagnosis and designed English prompts using the Pydantic schema to request precise output in JSON format. The prompt explicitly included criteria for each interpretation item, such as:

*"For the cyst in the image, please determine transparency, borders, internal structure, etc., and output the results in the specified JSON format."*

The prompt also clearly stated that the task was a *"professional interpretation task for dental radiology diagnosis,"* guiding GPT-4o to recognize the medical context before outputting.

**2) Prompts for Generating Findings**

When generating findings from the structured data, we used Japanese prompts to ensure clinically usable language, such as:

*"Based on the following structured data, please output dental radiology findings in a natural, medically sounding format."*

The output format followed the ground-truth style created by dental radiologists, naturally expressing location (e.g., "from no. 45 to no. 47") and characteristic descriptions (e.g., "a clearly demarcated unilocular translucency").

### 3) Output Format

Structured data were output in English using a unified JSON format with values stored as strings or lists. In contrast, the findings were output as natural Japanese sentences because the ground truth was prepared in Japanese. This style is concise and clinically applicable, with sentences ending in a period as the standard.

## 3. Results

### 3.1 Overall Performance Comparison

**Table 2** summarizes the performance of the two methods for each interpretation category, showing the mean correct answer rate for each item, together with the absolute and relative differences between the SLSO and CoT methods and the corresponding p-values.

Compared with the CoT method, the structured output method yielded higher accuracy in several interpretation items, most notably tooth number (66.9%), tooth displacement (33.3%), root resorption (28.6%), and relationship with other structures (9.0%). In contrast, clear improvements were not observed for radiolucency, internal structure, and boundary. For these items, correct answer rates were already high (0.864–1.000) for both methods, resulting in similar levels of accuracy.

Statistical tests were performed using the Shapiro-Wilk test to assess normality, followed by paired tests. A p-value $< 0.05$ was considered statistically significant, and $0.05 \leq p < 0.10$ was regarded as a trend toward significance. Owing to the limited sample size (22 cases), statistical significance was not achieved for any of the evaluation items.

An item-by-item analysis of Table 2 yielded the following three observations. For high-accuracy items (mean agreement rate > 0.8), namely radiolucency, internal structure, and boundary, both methods showed similarly high performance. For moderate-accuracy items (mean agreement rate 0.4–0.8), namely root resorption and relationship with other structures, the SLSO method consistently showed improvement. For tooth number identification, a low-accuracy item (mean agreement rate < 0.4), the structured method demonstrated a 66.9% relative improvement. This pattern is consistent with the contribution of the consistency-checking mechanism incorporated in the SLSO framework.

**Table 2** Comparison of correct answer rates between the CoT and SLSO methods across interpretation categories.

Mean correct answer rates (mean ± SE) for each interpretation category are shown for both methods, together with the absolute and relative differences between the two methods and the corresponding p-values.

| Interpretation category | CoT method (mean ± SE) | SLSO method (mean ± SE) | Absolute difference | Relative difference (%) | p-value |
|---|---|---|---|---|---|
| **Radiolucency** | 1.000 ± 0.000 | 1.000 ± 0.000 | +0.000 | +0.0 % | n/a |
| **Internal structure** | 0.909 ± 0.063 | 0.909 ± 0.063 | +0.000 | +0.0 % | n/a |
| **Boundary** | 0.864 ± 0.075 | 0.864 ± 0.075 | +0.000 | +0.0 % | n/a |
| **Root resorption** | 0.318 ± 0.102 | 0.409 ± 0.107 | +0.091 | +28.6 % | 0.248 |
| **Tooth displacement** | 0.273 ± 0.097 | 0.364 ± 0.105 | +0.091 | +33.3 % | 0.248 |
| **Relationship with other structures** | 0.500 ± 0.109 | 0.545 ± 0.109 | +0.045 | +9.0 % | 0.564 |
| **Tooth number** | 0.136 ± 0.075 | 0.227 ± 0.091 | +0.091 | +66.9 % | 0.180 |

Note:

- SE: Standard error
- Absolute difference = mean score of SLSO method - mean score of CoT method
- Relative difference = (absolute difference / mean score of CoT method) × 100 (%)

### 3.2 Comparison in Representative Cases

In this section, we provide a detailed analysis of representative cases classified as successful and unsuccessful examples to illustrate the effectiveness and limitations of the proposed method. A successful case was defined as one in which the structured output method achieved a substantial improvement in the average score for each interpretation item compared to the CoT method. Conversely, an unsuccessful case was defined as one in which the SLSO method showed a decrease in the average score for each interpretation item compared with the CoT method. For each case, the following five aspects were evaluated: (1) average score by interpretation item; (2) accuracy of tooth number identification; (3) conciseness and clinical appropriateness of the descriptions; (4) number of regeneration cycles (reflecting the effect of the self-correction loop); and (5) main improvements or common issues.

**Successful Case:**

**Figure 2** shows the input image for a successful case. **Table 3** lists the ground-truth finding, the report generated by the CoT method, the report generated by the SLSO method, the item-by-item scores for the two methods, the main improvements, and the number of regeneration cycles.

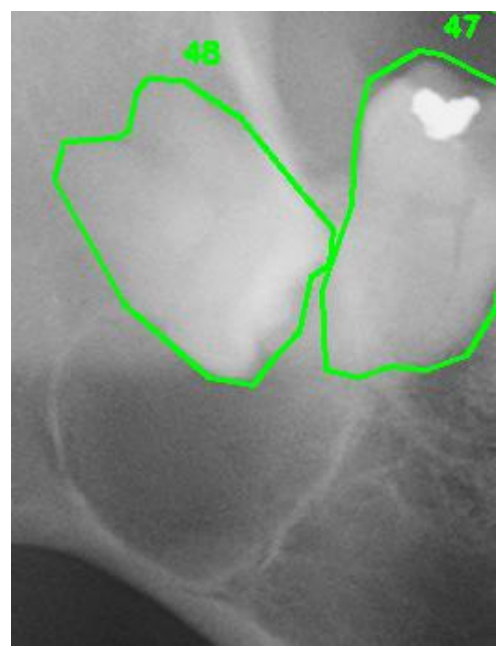

**Figure** 2 ROI image around cyst used as input for the successful case.

**Table 3** Generated outputs and evaluation results for the successful case

| **Category** | **Content** |
|---|---|
| **Ground-truth Findings** | The lower-right 8th tooth is horizontally impacted, with a well-defined, round radiolucent lesion surrounding the crown. The lesion extends to the inferior border cortical bone of the mandible. The mandibular canal appears to overlap with the lesion. A radiolucent area is observed in the crown of the lower right 7th tooth. The root is in contact with the lesion but no resorption is observed. Lower right 8th tooth: suspected dentigerous cyst. |
| **CoT-Method Output** | A well-defined, unilocular radiolucent lesion is observed in the mandibular molar region. The lesion presents a smooth contour, and radiographic findings suggest a cystic lesion (likely cyst). No obvious resorption is observed in the roots of teeth #47 and #48 (root resorption: none), with slight mesial displacement of the roots (tooth displacement: mild). Expansion of |
| | the inferior border cortical bone (cortical bone expansion: present) is noted, with bone expansion observed. The lesion overlaps with the inferior alveolar canal but shows no obvious invasion. There is no relationship with the maxillary sinus (unrelated), and no association with this anatomical structure is recognized. The teeth involved in the lesion are #47 and #48 in FDI notation. |
| **CoT Category-wise Score** | 0.71 |
| **SLSO-Method Output** | A radiolucent lesion is confirmed around teeth #47 and #48 on the right side of the mandible. The internal structure of the lesion is unilocular, with well-defined and round borders. These characteristics suggest a lesion with high probability of being a cyst. Although the lesion is adjacent to the inferior alveolar nerve canal, no pathological effects such as root resorption or tooth displacement are observed, and no cortical bone expansion is seen. Additionally, no association with the maxillary sinus is identified. |
| **SLSO-Method Category-wise Score** | 1.00 |
| **Main observations** | ✓ Accurate identification of tooth numbers ("teeth #47 and #48")<br>✓ Conciseness and clinical appropriateness of description<br>✓ Consistent description across all categories |
| **Number of Regenerations** | 5 times (automatic correction via tooth number consistency check) |

**Failure Case:**

**Figure 3** presents the input image for a failure case, and **Table 4** provides a comparison of the reports and evaluation outcomes for this case.

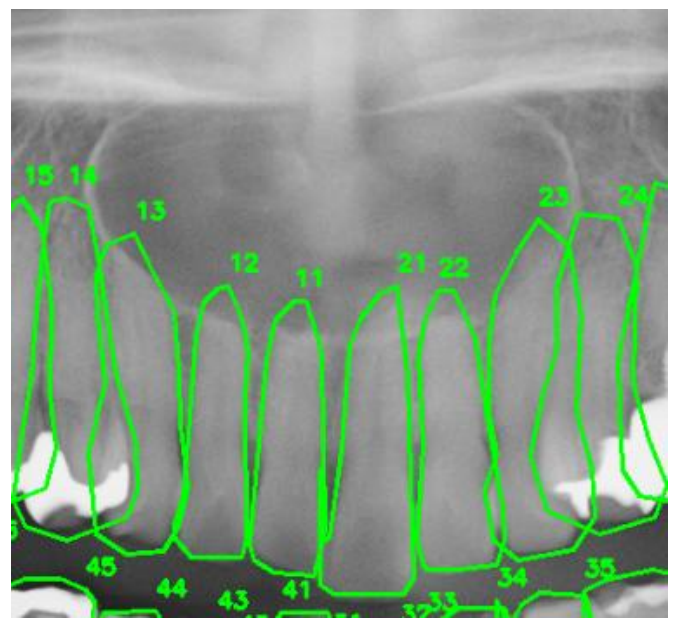

**Figure 3** ROI image around cyst used as input in the failure case.

**Table 4** Generated results and evaluation outcomes for the failure case.

| **Category** | **Content** |
|---|---|
| **Ground-truth Findings** | A cystic lesion is observed in the maxillary anterior region (from upper right 3 to upper left 3 apical area). It partially overlaps with the root apices, and mild resorption is also suspected. No left-right asymmetry is observed, but elevation of the nasal cavity floor is also suspected. No obvious calcifications or foreign bodies are observed internally. Suspected nasopalatine duct cyst. |
| **CoT-Method Output** | Step 1: Internal structure - The observed lesion shows radiolucency compared to surrounding structures, with high X-ray transparency. The morphology is unilocular, presenting a uniform radiolucent appearance.<br>Step 2: Lesion characteristics - The lesion boundary is well-defined, with a round to oval shape. Such findings suggest a suspected cyst (likely cyst).<br>Step 3: Effects on surrounding anatomical structures - No root resorption is observed in |
| | adjacent teeth (none). Tooth displacement is none to mild, with possible slight mobility particularly around tooth #21. No cortical bone expansion is observed (false).<br>Step 4: Relationship with anatomical structures - The lesion has no relationship with the inferior alveolar canal or maxillary sinus (unrelated).<br>Step 5: Involved tooth numbers (FDI system) - The lesion is located around tooth #21, with possible mild extension to teeth #11 and #22, but structural involvement is minimal. Therefore, the involved FDI tooth number is #21. |
| **CoT Category-wise Score** | 0.57 |
| **SLSO-Method Output** | A radiolucent lesion is observed around teeth #11 and #12 in the maxillary anterior region. The internal structure is unilocular, presenting a well-defined, round morphology. Based on these radiological findings, it is judged to have a high probability of being a cystic lesion. Regarding this lesion, no effects such as root resorption, tooth displacement, or cortical bone expansion were observed. Additionally, no relationship with the inferior alveolar canal or maxillary sinus is identified. |
| **SLSO-Method Category-wise Score** | 0.43 |
| **Main observations** | ✓ Difficulty in accurately identifying the extent of lesions spanning multiple teeth.<br>✓ Difficulty in evaluating relationships with anatomical structures (nasal cavity floor, incisive canal).<br>✓ Difficulty in detecting subtle findings (mild resorption). |
| **Number of Regenerations** | Once (automatic correction via tooth number consistency check). |

# 4. Discussion

This study demonstrated that the proposed SLSO framework improved the consistency and reliability of AI-generated radiological findings for jaw cysts in dental panoramic radiographs.

## 4.1 Effectiveness of Proposed Method

As noted in the Introduction, Liu et al.[10] reported substantial performance gaps between detection of radiopaque structures and radiolucent lesions using GPT-4o. Within this context, the present study explored whether a structured integrated processing framework could mitigate known weaknesses of current multimodal models. Given the known instability of direct structured-output prompting in GPT-4o due to medical refusal behavior, the Chain-of-Thought (CoT) method was selected as a realistic and reproducible baseline for comparison.

The proposed SLSO framework demonstrated consistent performance differences compared with the conventional CoT method in the generation of structured findings for jaw cysts on dental panoramic radiographs. Three characteristic technical effects were observed:

### 1) Improved Tooth Number Accuracy

The most notable difference between the two methods was observed in tooth number accuracy, which increased from 0.136 with the CoT method to 0.227 with the proposed method (+66.9%). This increase suggested a favorable trend associated with the consistency-checking mechanism at the core of the proposed framework. Unlike the CoT approach, which often yielded vague location expressions (e.g., "lower left mandibular molar region"), the structured output method enabled more precise identification (e.g., "tooth #47 and #48"). This effect could be attributed to the self-correction process implemented in Steps 4 (tooth number extraction) and 5 (consistency check). In one representative case, a perfect score of 1.0 was achieved after five regeneration cycles, illustrating the potential value of iterative correction. Although overall accuracy remained limited, these observations suggested that structured, self-correcting approaches might contribute to more reliable localization in dental radiographs.

### 2) Enforced Negative Findings and Improved Comprehensiveness

The structured schema required explicit "present/absent" judgments for each interpretation category, which promoted documentation of negative findings. This addressed a key weakness of the CoT approach, which frequently included vague expressions (e.g., "appears to…" or "may be…") and often omitted negative findings (e.g., "no evidence of resorption"). In contrast, the proposed framework enabled explicit and reproducible documentation of negative findings, such as "No evidence of pathological effects such as root resorption or tooth displacement." This mechanism resulted in higher completeness and internal consistency of the generated outputs, which was reflected in improved scores for root resorption, tooth displacement, and relationships with other structures.

**3) Hallucination Suppression Effect**

Constraints imposed by the structured format reduced references to nonexistent anatomical structures and logically inconsistent statements. From a technical perspective, this effect reflected the role of schema-based constraints in limiting the generative freedom of LLMs and enforcing clinically valid output spaces. For example, verbose CoT expressions such as "presents a smooth contour" were replaced with clinically appropriate descriptions such as "The internal structure of the lesion is unilocular, with a well-defined, round boundary."

These findings aligned with and extended the observations by Liu et al.[10]. While their study evaluated GPT-4o's detection capabilities using different tooth numbering systems, the present work addressed the complementary challenge of generating internally consistent and verifiable structured findings for radiolucent lesions. Importantly, the observed effects were achieved through an external processing framework rather than through modifications of the underlying vision-language model.

A distinguishing feature of the proposed approach was its design as an integrated processing framework rather than a model modification strategy. By implementing structured output constraints using Pydantic schemas and multi-stage validation loops external to the model, the SLSO framework represented a model-agnostic technical strategy for improving the re-liability of generative outputs in radiological reporting.

For example, as shown in Table 3, the tooth number consistency check detected a discrepancy between the tooth number item in the structured output and the tooth number extracted directly from the image, triggering five automatic regeneration cycles and ultimately correcting the hallucinated description to achieve a perfect category-wise score of 1.00. In contrast, as shown in Table 4, although one regeneration cycle was triggered by the same mechanism, the correction remained incomplete due to the complexity of the lesion spanning multiple teeth.

### 4.2 Study Limitations and Challenges

This study has several intrinsic limitations that should be acknowledged. This section addresses both study-level limitations and model/method-level limitations of the proposed framework.

**1) Dataset Size and Generalizability**

As a preliminary investigation, the dataset comprised only 22 cases from a single institution, limiting its statistical power and external validity. The limited availability of jaw cyst cases with confirmed histopathological diagnoses and expert radiologist annotations constrained further expansion at this stage. Despite these constraints, the dataset provided preliminary evidence suggesting potential improvements through the SLSO framework and demonstrated the technical feasibility of this integrated processing approach for future validation studies. Verification across multiple institutions, imaging devices, and patient populations would be essential to confirm the generalizability and robustness of the proposed approach in larger-scale studies. In addition, because annotated ROI images were used as inputs in this study, the applicability of the framework to full panoramic radiographs without manual localization remains an important topic for future investigation. Future studies may integrate automated lesion detection or tooth localization models with the proposed framework to enable a fully automated interpretation pipeline. Several deep learning-based approaches have been proposed for automated jaw cyst detection[16,17] and segmentation[18,19] in dental radiographs, as well as for tooth number identification[20], and integration with such methods represents a promising direction for future work.

**2) Complex Anatomical and Pathological Cases**

In the case of maxillary anterior failure (Table 4), the structured approach underperformed compared with CoT (0.43 vs. 0.57). This case involved extensive lesions spanning multiple teeth, complex anatomical relationships (e.g., nasal cavity floor and incisive canal), and subtle changes (e.g., mild resorption). Structural schema options (e.g., unilocular/multilocular) were insufficient to capture this complexity, and rigid structuring sometimes hindered appropriate descriptions. Future work will involve larger-scale analysis of failure patterns using expanded datasets to better characterize systematic sources of error, such as tooth number misidentification, difficulties in interpreting lesions spanning multiple teeth, and challenges in detecting subtle anatomical changes. Future improvements may involve integrating the proposed framework with models that provide stronger spatial localization capabilities. For example, combining the VLM-based interpretation framework with conventional deep-learning–based lesion detection or tooth localization models, or replacing the underlying VLM with newer models that offer improved spatial reasoning, may help ad-dress these limitations.

**3) Limits of Visual Recognition**

For persistently low-scoring items, such as tooth number (0.227) and tooth displacement (0.364), the inherent visual recognition capabilities of the GPT-4o were considered a limiting factor. In one failure case involving an extensive lesion spanning multiple teeth (Figure 3, Table 4), the AI output produced a more restricted range ("around teeth #11 and #12 in the maxillary anterior") compared to the ground truth ("apical region from right maxillary canine #3 to left maxillary canine #3"). While mild root resorption was detectable in other cases, neither method described subtle changes such as "mild resorption is also suspected" or "nasal cavity floor elevation is also suspected." These findings suggested that, in cases involving multiple complex anatomical structures, detecting subtle individual changes became even more challenging.

## 4.3 Experimental Technical Constraints

In addition to the study-level limitations, several technical issues arose during the comparative experiments.

**1) Difficulty in Direct Structured Output Generation**

In the initial phase of this study, we aimed to directly compare the proposed method with a simple approach that generates the same structured output format without a self-correction loop. However, unexpected technical challenges were encountered during these experiments. When GPT-4o was prompted to produce structured outputs directly (e.g., "Evaluate the cyst and output in JSON format"), it frequently generated refusal responses (e.g., "I cannot provide a diagnosis because I am not a doctor"), making it difficult to obtain stable comparative data. This behavior appeared to stem from GPT-4o's safety mechanisms that automatically reject medical diagnosis requests. Although reframing the context (e.g., "professional dental radiology interpretation task") or replacing "diagnosis" with "image analysis" could reduce refusals, such adjustments deviated from the notion of a "simple" structured output baseline.

**2) Choice of Comparison Method**

For the reasons stated above, the CoT method was adopted for comparison. CoT was a well-established approach to medical AI and its contrast to the SLSO framework offered valuable insights. Future research should explore minimally modified strategies to circumvent refusal errors and conduct more direct comparative experiments. This would enable a clearer separation of the effects of the structured output from the additional benefits of the self-correction loop.

**3) Evaluation Method**

The binary 0/1 scoring scheme was adopted to ensure objective and reproducible comparison by focusing on exact semantic consistency with ground-truth findings. However, this approach may underestimate partially correct outputs or clinically acceptable interpretations. Future studies may incorporate more nuanced evaluation approaches, such as semantic similarity metrics and multi-expert clinical scoring, to better reflect the practical utility of AI-generated radiological interpretations.

## 4.4 Implications for Clinical Application and Future Perspectives

**1) Role in Collaboration with Specialists**

This study suggested that AI-assisted diagnostic systems should function as supportive tools rather than as a replacement for specialists. Final confirmation by experts remained essential, particularly for complex anatomical interpretations, subtle or rare findings, and judgments requiring a clinical context. A stepwise introduction was considered effective, starting with support for clear and typical cases, moving to preliminary reading assistance under specialist confirmation, and eventually extending to educational use such as training residents. The structured nature of the proposed framework might facilitate such collaboration by providing transparent and verifiable intermediate outputs that could be reviewed by clinicians. Although the current performance remains limited for some categories, the proposed framework demonstrates potential for improving the reliability and consistency of AI-assisted radiological reporting rather than replacing expert interpretation.

**2) Study Comparison and Originality**

This preliminary study was among the first to address cyst diagnostic support using a multimodal approach with direct image input, extending beyond text-based approaches reported previously[12]. Rather than improving model performance through fine-tuning or architectural modification, the present work emphasized a framework-level contribution that complemented existing hallucination-suppression strategies such as RAG and CoT. Although the iterative design shared similarities with the Self-Refine framework[16], the inclusion of structured consistency checks distinguished the proposed method and aligned it with the strict requirements of medical imaging. Unlike the Self-Refine framework[21], which performs iterative refinement primarily at the text level, the proposed SLSO framework introduces structured intermediate representations that integrate image interpretation, structured data generation, and natural language findings, enabling consistency checks across different representation formats.

**3) Future Technical Improvements and Expansion**

Future work should involve larger datasets, specialist-in-the-loop evaluations, and semantic-level metrics to better capture clinical utility. Beyond jaw cysts, the proposed framework could be extended to other dental diseases and broader medical imaging domains, particularly when structured outputs could be aligned with domain-specific diagnostic criteria. The remaining challenges included reducing costs and processing time, and integration into clinical workflows.

**4) Data Privacy and Deployment Considerations**

In the present study, GPT-4o was accessed through a cloud-based API environment. For actual clinical deployment, however, the handling of medical images through external cloud services raises important issues related to patient privacy, data governance, and regulatory compliance. In many jurisdictions, the transmission of medical images to external servers may fall under regulations related to the protection of personal medical information, and strict institutional policies are often required.

In addition, in maxillofacial imaging, anatomical structures themselves may potentially serve as biometric identifiers, making complete anonymization challenging. There-fore, future clinical implementations will likely require locally deployable vision–language models operating within secure on-premises environments inside medical institutions. Such architectures would enable inference processing without transmitting medical images outside institutional networks, thereby addressing privacy and legal concerns.

The present study focused on methodological feasibility rather than clinical deployment; however, these regulatory and privacy considerations will be important topics for future research and system development.

**5) Model-Agnostic Design and Robustness to Future Model Evolution**

Although GPT-4o was used in the present study, the primary objective was not to evaluate a specific model version but to demonstrate a generalizable framework for improving the reliability of AI-generated radiological findings. Handler et al. recently discussed in Nature Medicine that even advanced generative models such as GPT-5 remained inherently fragile in medical applications, lacking intrinsic mechanisms for error detection and self-verification[22]. These limitations were unlikely to be resolved by model scaling alone in the near term. In this context, the proposed SLSO framework provided a practical pathway for enhancing robustness independently of model scaling and might serve as a foundation for future agent-based AI systems incorporating multiple reasoning and verification components.

## 5. Conclusions

This preliminary study demonstrated that the novel integrated processing approach, the Self-correction Loop with Structured Output (SLSO) framework, could achieve a modest yet meaningful improvement in the practical utility of GPT-4o for dental image interpretation. By enforcing structured documentation and incorporating an external validation mechanism, the proposed framework showed promising trends in improving the re-liability of AI-generated outputs, particularly in the explicit documentation of negative findings and the suppression of hallucinations. The stepwise and iterative nature of the proposed framework enables agent-like processing behaviors, while maintaining a hu-man-in-the-loop design suitable for clinical applications. Because the SLSO framework functions independently of model architecture, its underlying principles are not limited to dental imaging and may have potential applicability to other medical imaging domains and multimodal diagnostic tasks. While further validation using larger and more diverse datasets is warranted, the present findings provide technical insights into framework-level strategies for developing more reliable AI-assisted radiological reporting systems.

## Acknowledgements

We thank the clinical staff at Asahi University Medical and Dental Center and the technical staff at Gifu University for their support. This work was partially supported by a collaborative research project between Gifu University and EyeTech Co., Ltd.

## Additional Information

Supplementary information accompanies this paper.
Below is the link to the electronic supplementary material.
Supplementary Note 1- Preliminary Experiments
Supplementary Note 2- Detailed Description of the Proposed Framework

# Supplementary Information

**Generating Findings for Jaw Cysts in Dental Panoramic Radiographs Using a GPT-Based VLM: A Preliminary Study on Building a Two-Stage Self-Correction Loop with Structured Output (SLSO) Framework**

Nanaka Hosokawa[1], Ryo Takahashi[2], Tomoya Kitano[3], Yukihiro Iida[3],
Chisako Muramatsu[4], Tatsuro Hayashi[2], Yuta Seino[5], Xiangrong Zhou[6], Takeshi Hara[6],
Akitoshi Katsumata[3], and Hiroshi Fujita[6]

[1] Graduate School of Natural Science and Technology, Gifu University, Gifu, Japan
[2] EyeTech Co., Ltd, Tokyo, Japan
[3] School of Dentistry, Asahi University, Mizuho, Japan
[4] Faculty of Data Science, Shiga University, Hikone, Japan
[5] Division for Oral Dental Informatics, The University of Osaka Dental Hospital, Osaka, Japan
[6] Faculty of Engineering, Gifu University, Gifu, Japan

# Supplementary Note 1 - Preliminary Experiments

This note describes the preliminary phase of the study design, presents exploratory experiments that motivated the development of the proposed method, and summarizes their outcomes.

### Pilot Experiments

To develop a dental-image diagnosis-support system using GPT-4o, we incrementally improved the input image data preprocessing stage and prompt design, clarifying the issues at each stage before arriving at the final method design. The following section describes the considerations of each of the four phases.

### Phase 1: Direct Findings Generation Using Entire Image

In the pilot experiments, we compared the use of entire dental panoramic radiographs [**Figure S1 (a)**; original image size: 1976 × 976 pixels] sized to 224 × 224, 512 × 512, and 1024 × 1024 pixels. For image resizing, we zero-padded above and below the shortest edge (976 pixels vertically) of the original image to create a 1976×1976-pixel square image, and then used bilinear interpolation to shrink the image to the target size at the same ratio in both width and height. This method minimizes the distortion of anatomical structures while maintaining aspect ratio.

For each image size, we evaluated GPT-4o's finding generation using a standard prompt: "Please write a radiological finding description for the cyst shown in this dental panoramic radiograph, including its location, boundary, internal structure, and impact on the surrounding area."

As a result, we concluded that a 512×512-pixel image was the optimal setting, which had the lowest rate of error responses such as "I'm not a doctor, so I can't answer." We then directly input a 512 × 512-pixel image into GPT-4o to generate a cyst finding.

**Specific example results:**

Ground truth: "A clearly demarcated, unilocular radiolucency is observed from teeth 45 to 47."

GPT-4o output: "There appears to be a radiolucent lesion in the left mandibular molar region, with a somewhat indistinct border, and possible effects on the adjacent teeth."

The results of Phase 1 identified the following issues:

1) Processing load issues due to large image size.

2) Due to the diverse anatomical structures contained throughout the image, GPT-4o was distracted by findings other than cysts (crown restorations, implants, etc.).

3) Tooth number could not be identified (ambiguous expressions such as "left mandibular molar area").

4) Frequent use of vague expressions that are difficult to evaluate, such as "somewhat," "seems to be," and "possibly".

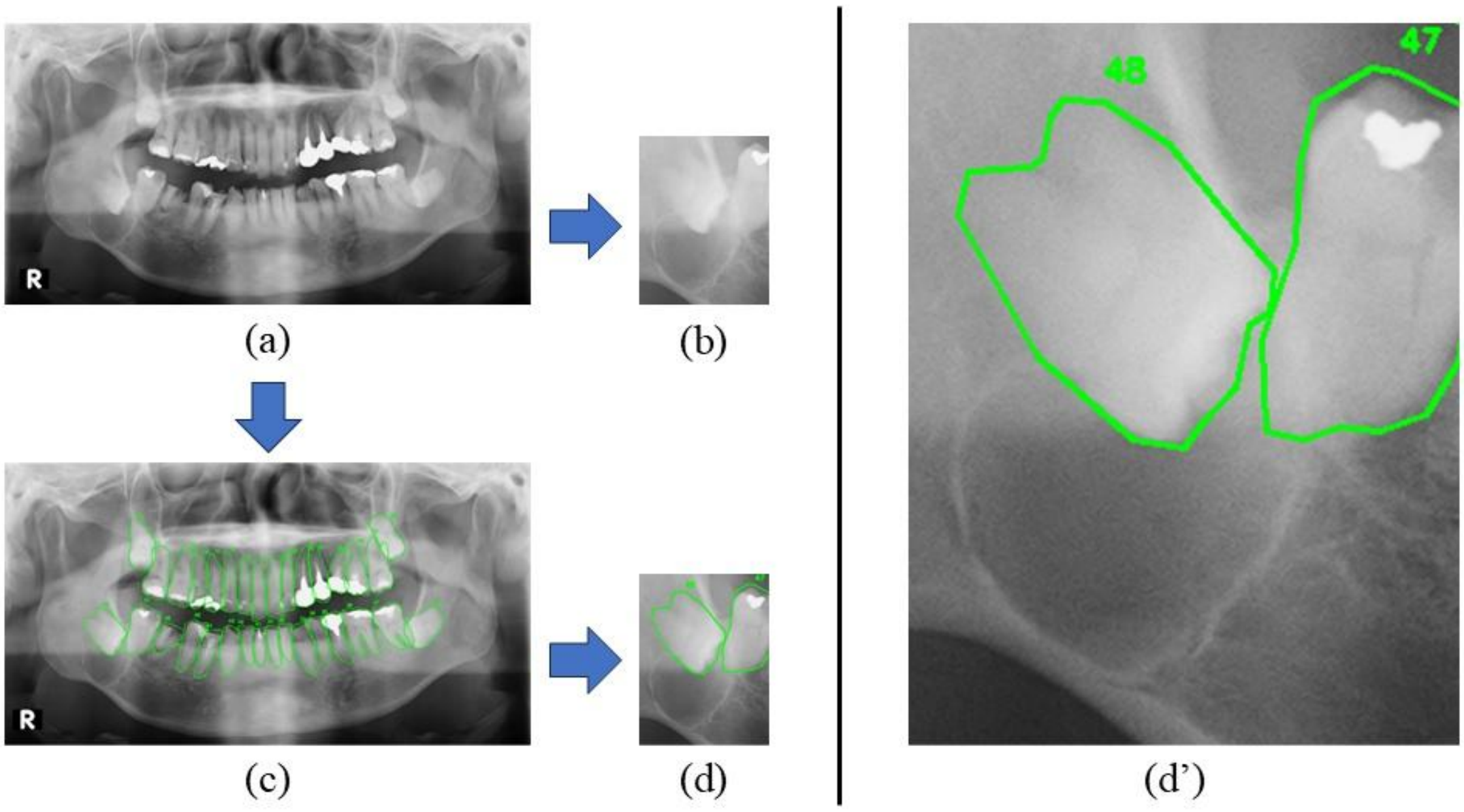

**Figure S1** Extraction of cyst regions from dental panoramic radiographs.

The preprocessing of panoramic radiographs is shown. (a) Original image (1976×976 pixels). (b) Simply cropped ROI image around the cyst (200-400 pixels, preserving the original resolution). (c) The original image in (a) with annotations (green outlines and FDI tooth numbers) of all tooth margins. (d) The cropped ROI image shown in (b), with annotated tooth margins and tooth numbers. (d’) Enlarged view to make the tooth numbers around the cyst more visible.

**Phase 2: Trial Improvements through Manual Extraction of ROIs**

In response to the challenges faced in Phase 1, we considered a method in which dentists manually extracted regions of interest (ROI) around the cysts [approximately 200–400 pixels; **Figure S1 (b)**] and fed the target area into the GPT-4o. While maintaining the basic structure of Phase 1, the verbal prompt was modified to reflect the characteristics of the ROI image: "Please describe in detail the location, boundaries, internal structure, and impact on surrounding tissues of the cyst in this extracted dental image region," encouraging precise observation within a limited area. While ROI extraction improved processing speed and significantly reduced the error rate compared to full image input, there were still many errors regarding cyst location and number of affected teeth (FDI number).

This manual ROI extraction process can be automated by incorporating a conventional deep learning-based cyst detection method. In recent years, research on the automated detection of cysts in dental panoramic radiographs has actively progressed.

Examples include an automated jaw cyst detection and classification system using a CNN architecture with an improved YOLOv3 architecture [16], a real-time cyst detection method using YOLOv2[17], and a segmentation method for odontogenic cyst lesions using U-Net-based multistage adaptive network[18]. An automated system for detecting the nasopalatine duct and apical cysts using EfficientDet-D3 achieved an mAP of 93.8%[19]. By integrating these technologies, the structured data generation framework proposed in this study is expected to be fully automated.

**Specific example results:**

Ground truth: "Clearly defined translucent regions are observed around no. 46 and no. 47."

GPT-4o output: "A radiolucent lesion is identified in the region of tooth #46, with a similar appearance observed adjacent to tooth #47."

The results of Phase 2 identified the following issues:

1) Although the accuracy of tooth number identification improved, ambiguous location expressions such as "vicinity" and "nearby" still remained.

2) It is difficult to grasp the exact positional relationship with adjacent teeth.

**Phase 3: Incorporation of Tooth Segmentation and Tooth Number Annotation**

To address the issue of positional relationship recognition in Phase 2, we considered an input format that combined the ROI images and affected tooth number information. Ideally, tooth segmentation and automatic FDI number assignment would be achieved using conventional deep learning techniques[20]. However, in this study, as a proof-of-concept, we verified the effectiveness of the proposed method using manual annotation by a dentist. **Figure S1 (c)** shows the original image annotated with tooth margins and tooth number, while **Figure S1 (d)** and **(d')** show the cropped ROI highlighting the cyst and affected teeth, with (d') providing an enlarged view for clarity. This representation is expected to provide the cyst and affected tooth information in a format that can be visually interpreted by the GPT model.

**Specific example results:**

Ground truth: "Clearly demarcated, unilocular translucencies are observed around teeth #46 and #47."

GPT-4o output: "A radiolucent area is observed around teeth #46 and #47, with relatively well-defined borders and showing mild to moderate

radiolucency."

The results of Phase 3 identified the following issues:
1) The accuracy of tooth number identification improved (correctly identifying "teeth #46 and #47").
2) Frequent use of ambiguous terms such as "relatively" and "mild to moderate" in descriptions of radiological features.
3) Occurrence of hallucination (unspecified content).

**Phase 4: Introducing CoT Prompts**

To mitigate the hallucination problem identified in Phase 3, we introduced CoT prompts that clearly explained the thought process step-by-step, and explored methods to encourage stepwise reasoning in GPT-4o. Specifically, each item in the schema defined in **Table 1** (radiographic transparency, internal structure, boundary, root resorption, tooth displacement, relationship to the mandibular canal and maxillary sinus, and affected tooth number, see Section 2.3 of the main text) was explicitly presented as an option, and the GPT-4o was instructed to simultaneously select an option for each item and write a final statement of findings within a single prompt.

**Example of a specific CoT prompt:**

"Analyze the cyst in this dental panoramic radiograph step by step using the following steps:
Step 1: Determine radiolucency (radiolucent/radiopaque).
Step 2: Evaluate the internal structure (unilocular/multilocular).
Step 3: Evaluate the clarity of the borders (well-defined/ill-defined).
Step 4: Evaluate the presence and extent of root resorption (no/mild/severe).
Step 5: Evaluate the presence and extent of tooth displacement (no/mild/severe).
Step 6: Evaluate the relationship with the mandibular canal and maxillary sinus (contact/displacement/invasion).
Step 7: Identify the affected tooth number (listed using FDI numbers).
Finally: Create a radiological statement based on the results of the above analysis."

Similar to Phase 3, the ROI images were provided with FDI tooth number annotations and margin information for each tooth. While this improved the transparency of the reasoning process, it did not fundamentally resolve the issues of ensuring the factuality of the final findings or the verifiability of the written content. The results of the comparison between this CoT method and the structured method described below are detailed in Section 3.

**Specific example results:**

Ground truth: "A clearly demarcated unilocular radiolucency is observed around teeth numbers 46 and 47."

GPT-4o output (CoT): "First, the location is assessed: a radiolucent area is observed in the mandibular molar region. Next, the border is evaluated: the margin appears to be well-defined, although not with certainty. The radiolucency is then judged: it seems to be of moderate degree. In conclusion, there is a possible cyst-like lesion around tooth #46."

The results of Phase 4 identified the following issues:

1) The transparency of the thought process was improved by stepwise reasoning.
2) Frequent use of uncertain expressions such as "seems," "appears," and "possibly."
3) It was difficult to guarantee the factuality of the final findings.
4) The issue of the verifiability of the written content was not fundamentally resolved.

## Supplementary Note 2 - Detailed Description of Proposed Framework

This supplementary note provides a detailed description of each step (1–10) of the proposed framework, including the representative prompts, schema definitions, and example outputs. These details complement the overview presented in the main text (Section 2.4, **Figure 1**).

---

**Step 1: Input**

ROI images annotated with tooth margins and FDI numbers surrounding the jaw cyst are input along with verbal prompts containing interpretation instructions.

**Example prompt:**

*"You are a professional dental radiologist. Analyze this dental panoramic radiograph and output the results as structured data according to the specified items."*

---

**Step 2: GPT-4o Image Analysis (Start of Parallel Processing)**

Utilizing the GPT-4o multimodal capabilities, the ROI image is analyzed. From this step onward, parallel processing is employed while simultaneously performing structured data generation (Step 3) and tooth number extraction (Step 4).

---

**Step 3: Structured Data Generation**

GPT-4o is instructed to generate structured data (in JSON format) according to the schema defined in **Table 1**. Specifically, the interpretation item schema was defined using the Pydantic library, and the radiological features of the lesion (jaw cyst) in the image were extracted as strictly structured data by setting the Pydantic schema in OpenAI API (Application Programming Interface), response format parameters.

**Pydantic schema definition example:**

```
class CystAnalysis(BaseModel):
    radiolucency: Literal["radiolucent", "radiopaque"]
    internal_structure: Literal["unilocular", "multilocular"]
    boundary: Literal["well-defined", "ill-defined"]
    root_resorption: Literal["no", "mild", "severe"]
    # ...
```

**Example output:**

```
{
 "radiolucency": "radiolucent",
 "internal_structure": "unilocular",
 "boundary": "well-defined",
 "root_resorption": "mild",
 "tooth_displacement": "no",
 "anatomical_relation": "contact",
 "affected_teeth": ["33", "34", "35", "36"]
}
```

---

**Step 4: Tooth Number Extraction**

In parallel, GPT-4o's visual recognition capabilities are used to extract textual tooth numbers from the input image for teeth suspected to be affected by jaw cysts. This information is obtained separately from the structured data and used for the subsequent consistency assessment.

---

**Step 5: Tooth Number Consistency Check**

The inclusion and consistency of the number of affected teeth described in the structured data obtained in Step 3 and the tooth number list extracted in Step 4 are evaluated. For example, if the structured data describes "33, 34" and the extracted results from the image are "33, 34, 35, 36," this is considered a match. However, if a complete mismatch exists, regeneration is required.

---

**Step 6: Regenerate Structured Data & Re-extract Tooth Numbers**

If an inconsistency is recognized in Step 5, GPT-4o is instructed to regenerate the structured data, as shown in the following example. In this case, specific feedback information regarding tooth number mismatch is created, including 1) the tooth number identified in the structured data, 2) the tooth number extracted directly from the image, and 3) a prompt clearly indicating the differences between the two, and the structured data is regenerated. Subsequently, the system returns to Step 5 and repeats the consistency check.

**Example prompt:**

*"An inconsistency in tooth numbers was detected in the previous structured data*

*generation.*

*Tooth numbers identified in the structured data: [31, 32]*

*Tooth numbers extracted directly from the image: [11, 12, 13, 21, 22, 23]*

*Difference: Numbers 11, 12, 13, 21, 22, and 23 are not included in the structured data. Please recheck the image and regenerate structured data in JSON format, including all teeth affected by the lesion."*

---

**Step 7: Finding Generation**

Based on the structured data, whose consistency was confirmed, GPT-4o is instructed to generate findings in a natural language format. In this case, the generated findings mimicked the actual radiographic findings written by dentists, aiming for quality that can be used directly in clinical reports.

**Example prompt:**

*"Based on the following structured data, please output a dental radiology finding sentence in natural-sounding Japanese medical description. Please standardize location expressions in the format of 'from number XX to number XX' and create a concise, clinically appropriate finding sentence that includes transparency, boundary, and internal structure. [Insert JSON of tooth number consistency-checked structured data]"*

---

**Step 8: Regenerate Structured Data from Findings**

The sentence generated in Step 7 is re-input into GPT-4o, and the information contained in the sentence is converted back into structured data. This enables the system to verify whether the natural language representation accurately reflects the original structured information.

**Example prompt:**

*"Parse the following dental radiological findings and extract the information contained therein as structured data (JSON format). Output the following items according to the defined schema: radiolucency, internal structure, boundary, root resorption, tooth displacement, anatomical relationship, and affected tooth number. [Insert the findings generated in Step 7]"*

---

**Step 9: Structured Data Consistency Check**

The consistency between the data restructured in Step 8 and those generated in Step

3 is verified for the following items:

- Matching of affected tooth numbers: Detection of missing tooth numbers.
- Match radiological findings such as position, translucency, internal structure (unilocular/multilocular), and boundary.
- Semantic discrepancies: identification of inconsistencies arising from ambiguous terms or synonymous substitutions.

If consistency is confirmed, the findings are produced as the final output.

---

**Step 10: Regeneration Findings**

If an inconsistency was detected in Step 9, GPT-4o was used to regenerate the findings. Structured data (consistency verified in Step 5 and corrected in Step 6 if necessary) are retained, while only the findings are regenerated, with feedback specifying the inconsistency (e.g., "The tooth number is missing" or "The boundary description differs from the structured data"). The regenerated findings are then reverified for consistency in Steps 8 and 9.

**Example prompt:**

*"An inconsistency issue was detected in the previously generated findings. Specifically, it is [content of the inconsistency]. Please regenerate the findings accurately based on the following structured data:*
*[Insert tooth number consistency-checked structured data]"*

---